\documentclass[letterpaper]{article} % DO NOT CHANGE THIS
\usepackage{aaai2026}  % DO NOT CHANGE THIS
\usepackage{times}  % DO NOT CHANGE THIS
\usepackage{helvet}  % DO NOT CHANGE THIS
\usepackage{courier}  % DO NOT CHANGE THIS
\usepackage[hyphens]{url}  % DO NOT CHANGE THIS
\usepackage{graphicx} % DO NOT CHANGE THIS
\usepackage{natbib}  % DO NOT CHANGE THIS AND DO NOT ADD ANY OPTIONS TO IT
\usepackage{caption} % DO NOT CHANGE THIS AND DO NOT ADD ANY OPTIONS TO IT
\usepackage{algorithm}
\usepackage{algorithmic}
\usepackage{amsmath}
\usepackage{booktabs} 
\usepackage{newfloat}
\usepackage{listings}
\DeclareCaptionStyle{ruled}{labelfont=normalfont,labelsep=colon,strut=off} % DO NOT CHANGE THIS
\floatstyle{ruled}
\newfloat{listing}{tb}{lst}{}
\floatname{listing}{Listing}
\title{MagicRoad: Semantic-Aware 3D Road Surface Reconstruction via Obstacle Inpainting}
\author{
    Xingyue Peng$^{1}$\equalcontrib, Yuandong Lyu$^{1}$\equalcontrib, Lang Zhang$^{1}$\thanks{Corresponding author}, Jian Zhu$^{1}$, Songtao Wang$^{1}$, Jiaxin Deng$^{1}$, Songxin Lu$^{1}$, Weiliang Ma$^{1}$, Dangen She$^{1}$, Peng Jia$^{1}$, XianPeng Lang$^{1}$
}
\affiliations{
    \textsuperscript{\rm 1}Li Auto Inc.\\
    
}

\usepackage{bibentry}
\usepackage{colortbl}
\usepackage{graphicx}
\usepackage{subcaption}
\usepackage{xspace}
\usepackage{multirow}
\usepackage{makecell}
\usepackage{algorithm}
\newcommand{\bc}{\mathbf{c}}

\newcommand{\bn}{\mathbf{n}}\newcommand{\bN}{\mathbf{N}}

\newcommand{\bp}{\mathbf{p}}

\newcommand{\bs}{\mathbf{s}}
\newcommand{\bt}{\mathbf{t}}
\newcommand{\bu}{\mathbf{u}}

\newcommand{\bx}{\mathbf{x}}

\newcommand{\bSigma}{\boldsymbol{\Sigma}}

\newcommand{\cG}{\mathcal{G}}

\makeatletter
\DeclareRobustCommand\onedot{\futurelet\@let@token\@onedot}
\def\@onedot{\ifx\@let@token.\else.\null\fi\xspace}
\def\eg{e.g\onedot}

\def\etal{et~al\onedot}

\makeatother

\definecolor{yellow}{rgb}{1, 1, 0.7}
\definecolor{orange}{rgb}{1, 0.85, 0.7}
\definecolor{tablered}{rgb}{1, 0.7, 0.7}
\definecolor{red}{rgb}{1, 0, 0}
\definecolor{wincolor}{rgb}{0.85, 0.0, 0.0}
\definecolor{darkyellow}{rgb}{0.8, 0.8, 0.5}
\definecolor{darkred}{rgb}{0.7, 0.3, 0.3}
\definecolor{darkgreen}{rgb}{0.3, 0.7, 0.3}
\definecolor{green}{rgb}{0, 1.0, 0}
\definecolor{pink}{rgb}{1, 0.4, 0.7}

\newcommand{\ourmethod}{MagicRoad}

\begin{document}

\maketitle

% \begin{abstract}
% AAAI creates proceedings, working notes, and technical reports directly from electronic source furnished by the authors. To ensure that all papers in the publication have a uniform appearance, authors must adhere to the following instructions.
% \end{abstract}


\begin{abstract}
Road surface reconstruction is essential for autonomous driving, supporting centimeter-accurate lane perception and high-definition mapping in complex urban environments.
While recent methods based on mesh rendering or 3D Gaussian splatting (3DGS) achieve promising results under clean and static conditions, they remain vulnerable to occlusions from dynamic agents, visual clutter from static obstacles, and appearance degradation caused by lighting and weather changes.
We present a robust reconstruction framework that integrates occlusion-aware 2D Gaussian surfels with semantic-guided color enhancement to recover clean, consistent road surfaces.
Our method leverages a planar-adapted Gaussian representation for efficient large-scale modeling, employs segmentation-guided video inpainting to remove both dynamic and static foreground objects, and enhances color coherence via semantic-aware correction in HSV space.
Extensive experiments on urban-scale datasets demonstrate that our framework produces visually coherent and geometrically faithful reconstructions, significantly outperforming prior methods under real-world conditions.
\end{abstract}

\section{Introduction}
\label{sec:intro}

Accurate and complete reconstruction of road surfaces is a fundamental task for a wide range of real-world applications, including autonomous driving, urban planning, and high-definition maps. 
The emergence of Bird's-Eye View (BEV) perception paradigms has revolutionized this domain by providing a unified spatial representation that bridges multi-view camera inputs with 3D geometric understanding.
By projecting the scene onto a ground-aligned plane, BEV simplifies downstream tasks such as lane detection, obstacle avoidance, and road topology estimation~\cite{xia2024dumapnet, xia2025ldmapnet}.

\begin{figure}[!thb] \centering
    \includegraphics[width=0.48\textwidth]{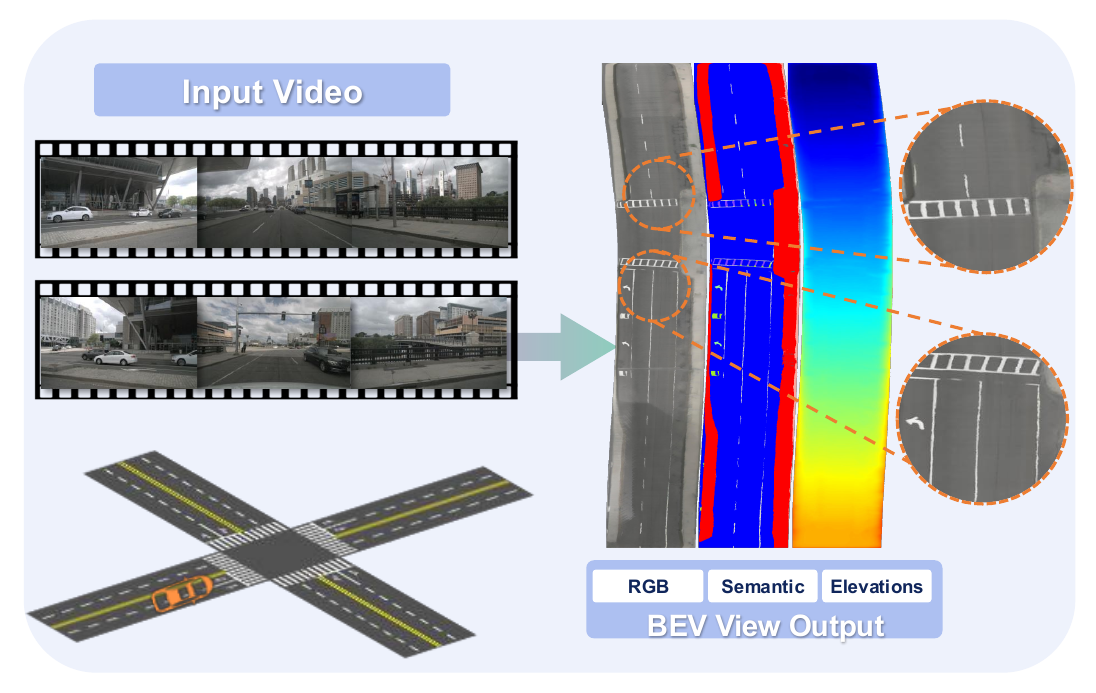}
    \caption{We introduce {\ourmethod}, an innovative framework for large-scale clean road surface reconstruction from in-car camera videos, which outputs BEV road maps with RGB appearance, semantic labels, and elevation information.} 
    \label{fig:figure1}
\end{figure}

Despite these advantages, achieving high-quality road surface reconstruction under the BEV framework remains a challenging and underexplored problem—yet it is crucial for producing reliable, interpretable, and structured maps in complex urban environments. The main difficulties arise from dynamic occlusions caused by moving vehicles and pedestrians, persistent visual clutter from static objects such as parked cars, and significant appearance variations due to lighting changes, shadows, and weather conditions.

Previous road surface reconstruction approaches are predominantly based on multi-view stereo (MVS)~\cite{hartley2003multiple,pollefeys2004visual,schonberger2016structure}, which relies on photometric consistency across images to estimate 3D geometry. While effective in well-textured, static scenes, MVS methods often break down in large scale urban environments.  
More recently, neural rendering techniques such as Neural Radiance Fields (NeRF)~\cite{mildenhall2021nerf, tancik2022block, turki2022mega} have shown impressive results in capturing view-dependent appearance and fine details. However, NeRF-based methods typically require dense multi-view observations, suffer from slow inference, and are not inherently designed for ground-plane-aligned reconstruction.
To address these issues, some works have explored explicit and compact representations like 3D Gaussians~\cite{kerbl20233d}, which provide efficient and scalable modeling of scene geometry and appearance. These methods enable faster rendering and easier integration with downstream tasks in BEV space. While promising, all of the above approaches share a common limitation: they assume the input scene is clean and static. None of them explicitly handle transient foreground objects such as moving pedestrians and vehicles, nor do they compensate for significant appearance variations caused by lighting changes, shadows, or occlusions. This often leads to noisy, inconsistent, or incomplete road surface reconstructions, especially in real-world outdoor settings.

To address these challenges, we propose {\ourmethod}, a unified framework that couples semantic-aware inpainting with adaptive appearance enhancement for robust large-scale road surface reconstruction. Given an input scene, our method first employs a segmentation-guided inpainting network to remove dynamic and static foreground objects, effectively recovering the occluded road textures. The cleaned image is then passed through a semantic-guided enhancement module, which corrects lighting-induced inconsistencies and improves visual coherence by leveraging semantic priors. These two stages—foreground removal and appearance harmonization—effectively restore structural and visual consistency across views, enabling more accurate and coherent reconstruction with our 2D Gaussian-based representation. This design enables {\ourmethod} to recover high-fidelity, semantically aligned road surfaces even in visually degraded or cluttered scenarios.

Extensive experiments on urban-scale driving datasets demonstrate that {\ourmethod} produces cleaner, more consistent, and geometrically faithful reconstructions compared to existing baselines, particularly in complex urban scenes. Our method bridges the gap between low-level surface modeling and high-level semantic understanding, providing a robust solution for road-centric 3D scene reconstruction.

Our contribution can be summarized as follows:
\begin{itemize}
\item We present {\ourmethod}, a unified framework for large-scale road surface reconstruction that integrates semantic-aware inpainting and color enhancement to handle dynamic occlusions and lighting variations.
\item We design a segmentation-guided inpainting pipeline that effectively removes both dynamic (\eg, moving vehicles) and static (\eg, parked cars) occluders from multi-view road images.
\item We introduce a semantic-aware color enhancement module that improves cross-view color consistency and highlights road structural elements such as lane markings and curbs by leveraging semantic priors in HSV color space.
\item Extensive experiments on real-world driving datasets show that {\ourmethod} achieves cleaner, more consistent, and more accurate reconstructions compared to state-of-the-art baselines.
\end{itemize}

\section{Related Works}
\label{sec:related}

\subsection{3D Reconstruction}

Traditional 3D reconstruction methods, such as Multi-View Stereo (MVS)~\cite{zhou2021dp, fan2018road, schonberger2016pixelwise}, recover dense depth maps by enforcing photometric consistency across calibrated views. While accurate in static, textured scenes, they struggle with dynamic content and view inconsistencies, common in outdoor driving scenarios.

Recent advances explore implicit representations, such as neural radiance fields (NeRFs)~\cite{mildenhall2021nerf, tancik2022block, turki2022mega}, which model continuous geometry and appearance via neural networks. These methods achieve high-fidelity synthesis but require dense views, are slow to train and render, and are ill-suited for structured planar surfaces like roads.

Gaussian-based methods~\cite{kerbl20233d, huang20242d} offer a middle ground, using compact, learnable surfels for fast rendering, occlusion handling, and BEV compatibility. However, most assume static, clean scenes and lack mechanisms to handle transient clutter or inconsistent lighting.

In road-specific domains, RoMe~\cite{mei2024rome} employs mesh-based rendering optimization, while RoGS~\cite{feng2024rogs} adopts meshgrid-aligned 3D Gaussians for efficient BEV reconstruction. Despite their strengths, both methods ignore dynamic occlusions and appearance variation, limiting their robustness in real-world conditions.

These limitations motivate our approach, which enhances Gaussian-based representations with semantic-aware inpainting and color harmonization, producing clean, consistent reconstructions in dynamic urban environments.

\begin{figure*}[!ht] 
    \centering
    \includegraphics[width=1.0\textwidth]{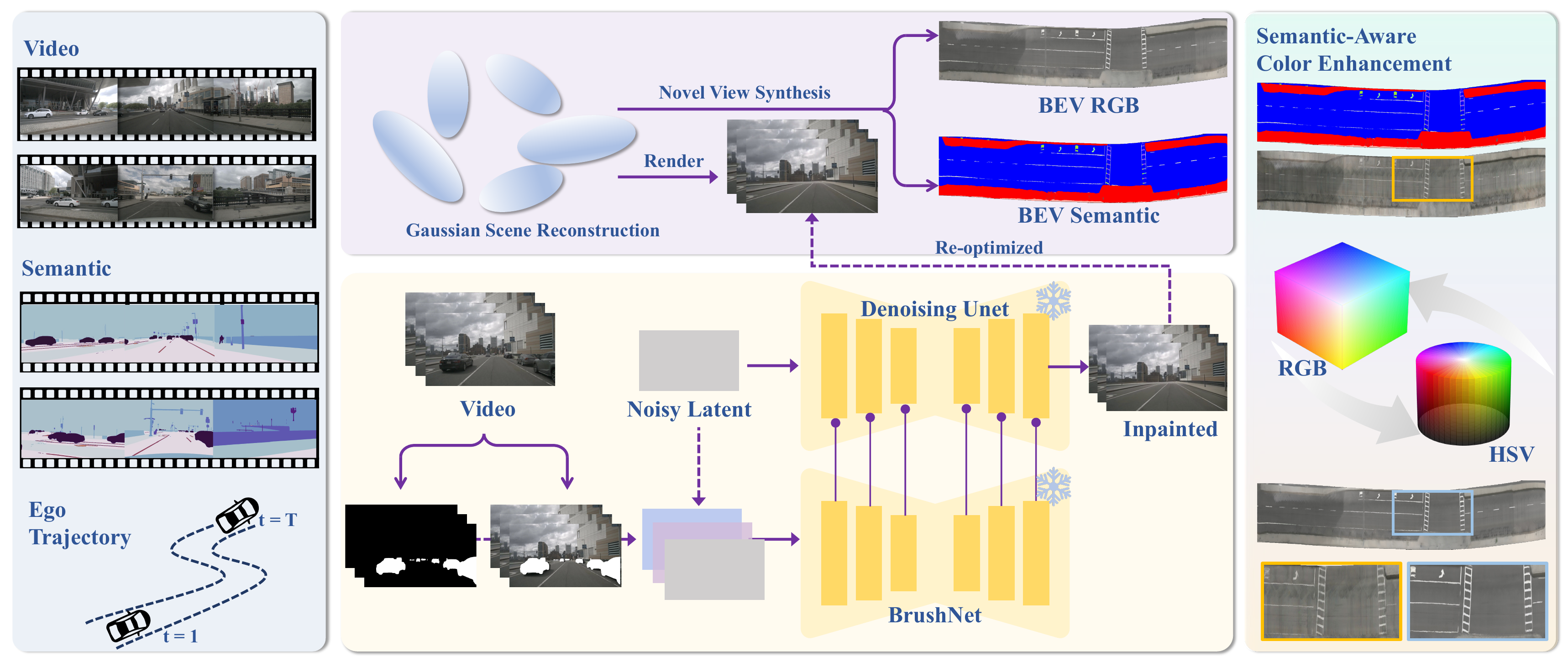}
    % example-grid-100x100pt
    % example-image-a
    \caption{Overview of \ourmethod. Given in-car camera videos, we perform temporal inpainting to remove dynamic objects, enhance appearance consistency via semantic-aware color enhancement, and reconstruct clean road surfaces using optimized 2D Gaussian surfels, enabling high-fidelity BEV rendering in both RGB and semantic space.} 
    \label{fig:figure2}
\end{figure*}

\begin{figure}[!tb] \centering
    \includegraphics[width=0.48\textwidth]{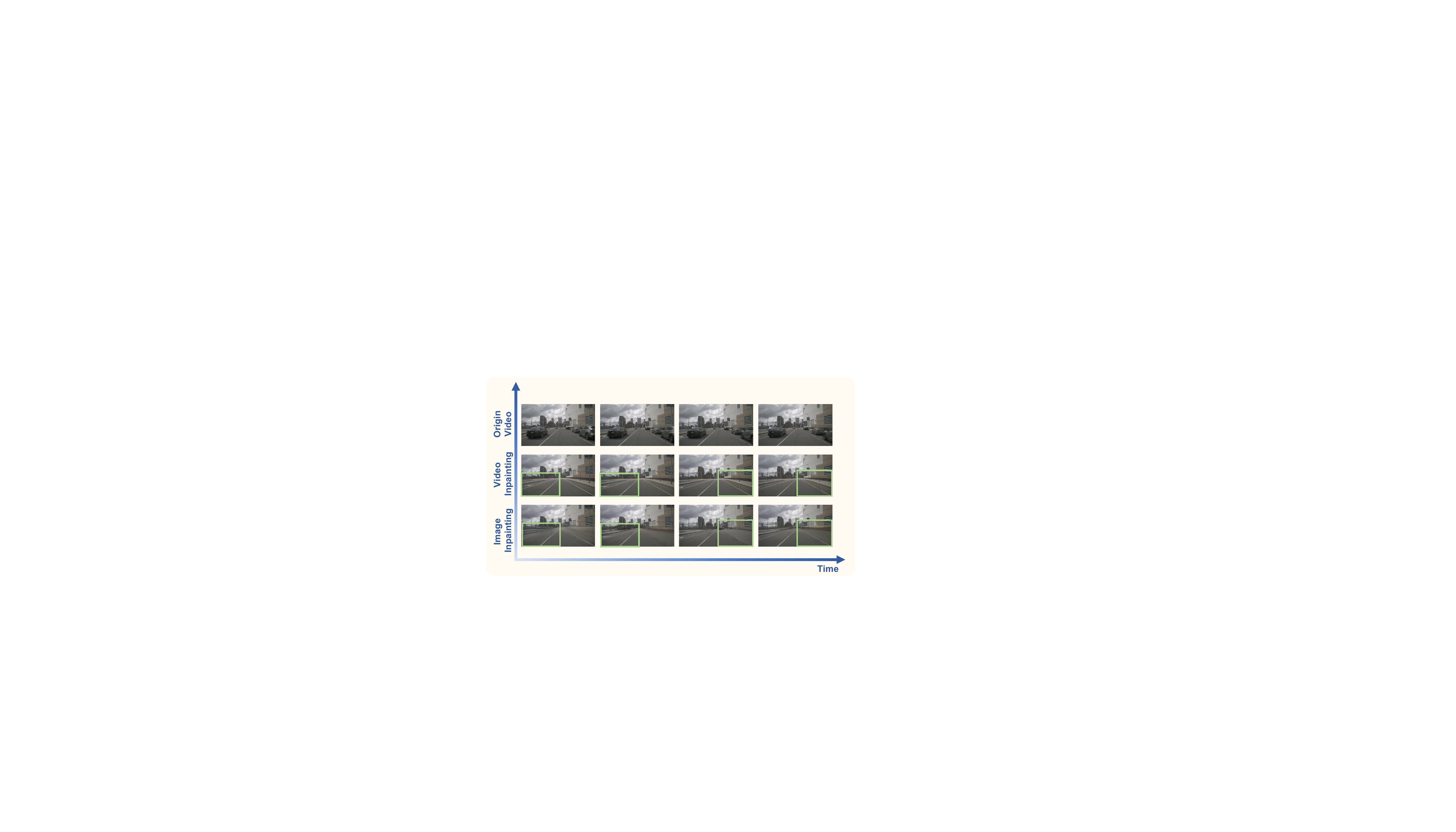}
    \caption{Video inpainting leverages temporal attention across frames, leading to more consistent and realistic reconstruction of occluded regions. Unlike image-based inpainting, it preserves continuity across time, resulting in smoother appearance and fewer artifacts in sequential frames.} 
    \label{fig:figure3}
\end{figure}

\subsection{Image and video inpainting.} 
Image and video inpainting~\cite{bertalmio2000image} aims to restore missing or occluded regions with plausible, context-consistent content. Early patch-based methods~\cite{huang2014image, guo2017patch} propagated textures from known areas, but struggled with complex semantics and structure.

Deep learning has significantly advanced inpainting. CNN-based approaches~\cite{yu2018generative, yu2019free} introduced contextual attention and gated convolutions to better model spatial context. GANs~\cite{pathak2016context, zhao2021large} improved realism via adversarial supervision, while Transformers~\cite{wan2021high, zhou2023propainter, cao2023zits++} and diffusion models~\cite{rombach2022high, saharia2022palette, lugmayr2022repaint} further enhanced global reasoning and visual fidelity. However, most of these methods target general natural images and lack explicit modeling for structured outdoor scenes.

Video inpainting extends the task to the temporal domain, requiring both spatial plausibility and temporal coherence. Early methods~\cite{wexler2007space, newson2014video} used patch volumes or trajectory tracking, but were limited by motion complexity. Deep models~\cite{kim2019deep, xu2019deep, chang2019vornet} introduced flow-guided or GAN-based supervision, improving temporal realism. More recent works leverage Transformers~\cite{ke2021occlusion} and diffusion models~\cite{zhang2022flow, ren2022dlformer} for long-range consistency, though many remain computationally demanding.

Despite these advances, existing inpainting methods are rarely tailored for structured, large-scale road scenes—where occlusions align with specific semantic categories and consistent surface restoration is critical. In contrast, we adopt a semantic-aware inpainting framework that explicitly leverages class-aware guidance and temporal information, producing clean, occlusion-free inputs to supervise downstream 3D reconstruction.
\section{Method}
\label{sec:method}

\begin{figure*}[!ht] 
    \centering
    \includegraphics[width=1.0\textwidth]{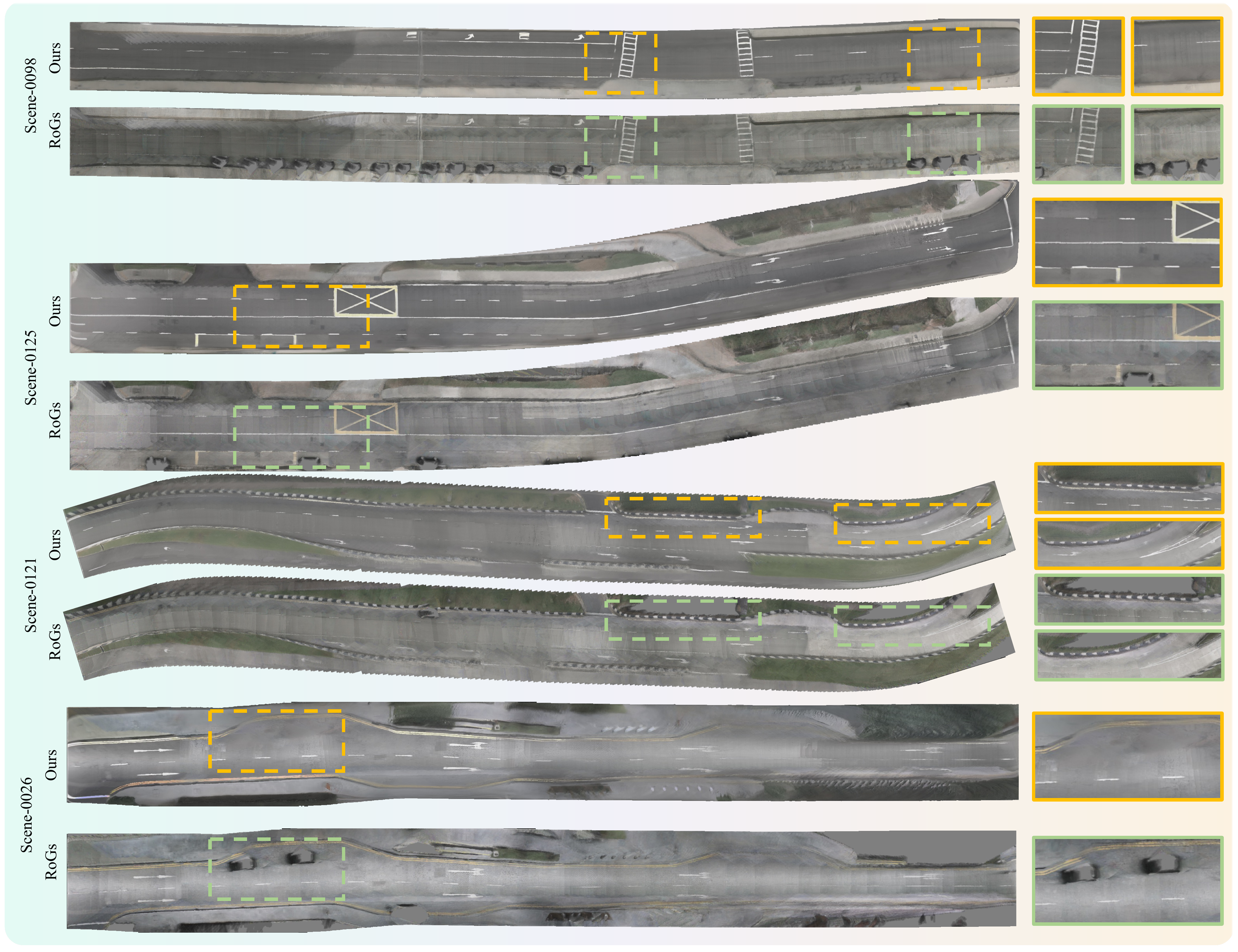}
    % example-grid-100x100pt
    % example-image-a
    \caption{Qualitative comparison results of our methods. Our method effectively fills reconstruction holes caused by parked vehicles, producing structurally complete and semantically consistent road surfaces. In particular, fine details such as parking lines are plausibly restored, and our approach achieves better color consistency and sharper lane markings compared to RoGS.} 
    \label{fig:figure4}
\end{figure*}

\subsection{Planar Gaussian Surfel Representation}
Prior works predominantly employ either explicit mesh-based representations or implicit neural radiance fields (NeRF) for scene reconstruction. 
While Kerbl~\etal~\cite{kerbl20233d} pioneered the use of 3D Gaussian Splatting (3DGS) with 3D covariance matrix $\bSigma$ and their location $\bp_k$, achieving remarkable results in general scene modeling, these volumetric primitives exhibit two critical limitations for road surface reconstruction: (1) Unnecessary degrees of freedom in the vertical dimension, and (2) Multi-view inconsistency caused by view-dependent depth ambiguity. 
Huang~\etal~\cite{huang20242d} subsequently demonstrated that constraining Gaussians to 2D manifolds improves geometric stability for planar structures.

\subsubsection{Preliminary.}
Building upon these insights, we adopt planar-adapted 2D Gaussian surfels as our fundamental representation, each surfel $\cG(\bu)$ is parameterized by:
\begin{equation}
\cG(\bu) = \exp\left(-\frac{u^2+v^2}{2}\right)
    \label{eq:gaussian-2d}
\end{equation}
where the point $\bu=(u,v)$ is defined in the tangent plane:
\begin{equation}
P(u,v) = \bp_k + s_u \bt_u u + s_v \bt_v v
\label{eq:plane-to-world}
\end{equation}
The center $\bp_k$, scaling $(s_u,s_v)$, and the rotation  $(\bt_u, \bt_v)$ are learnable parameters. Gaussians are sorted by their depth values along each viewing ray, enabling correct occlusion handling. The final pixel color $\bc(\bx)$ is computed through front-to-back blending:
\begin{equation}
\bc(\bx) = \sum_{i=1} \bc_i\,\alpha_i\,\cG_i(\bu(\bx)) \prod_{j=1}^{i-1} (1 - \alpha_j\,\cG_j(\bu(\bx)))
\end{equation}
where $\alpha_i \in [0,1]$ controls opacity and $\mathbf{c}_i$ represents view-dependent color modeled via third-order spherical harmonics.

\subsubsection{Planar Gaussian for Road Reconstruction.}
To enhance the model's semantic perception capability and support subsequent semantic-guided color enhancement, inspired from~\cite{zhou2024hugs, zhou2024feature, yan2024street}, we extend the traditional Gaussian splatting representation with semantic information injection. Each Gaussian surfel is augmented with a semantic feature descriptor in addition to standard geometric and appearance attributes. A planar gaussian surfel can be parameterized explicitly as:
\begin{equation}
  \Theta = \{\mathbf{p}_k, s_u, s_v, \mathbf{t}_u, \mathbf{t}_v, \alpha, \bc, \bs\},
\end{equation}
where $\alpha$ represents opacity, $\bc$ represents the color, and we add the attribute $\bs$ as semantic information.

\begin{figure*}[!ht] 
    \centering
    \includegraphics[width=1.0\textwidth]{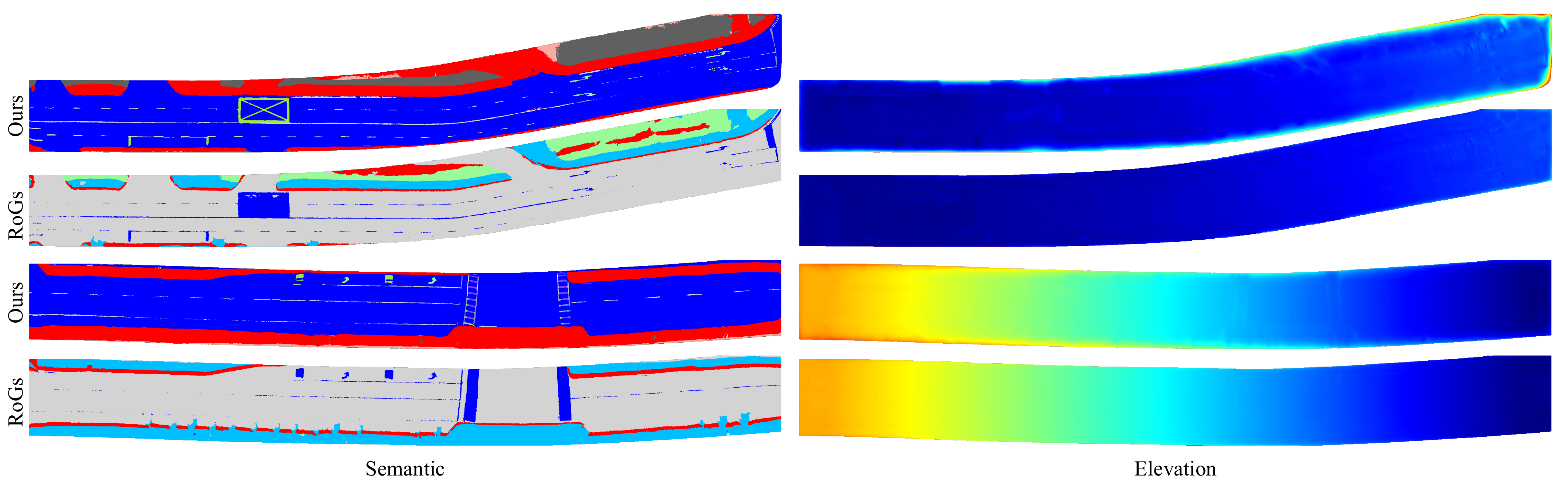}
    % example-grid-100x100pt
    % example-image-a
    \caption{Semantic and elevation comparison. Our method recovers richer semantic details such as fine-grained lane markings. In the elevation map, compared to the LiDAR-free version of RoGS, our approach better reconstructs structural elements like road edges and curbs, despite being entirely LiDAR-free.} 
    \label{fig:figure5}
\end{figure*}

\subsection{Obstacle Inpainting}

To address transient occlusions—such as moving vehicles and pedestrians in multi-view road surface reconstruction, we propose a video inpainting framework tailored to the unique challenges of autonomous driving. Unlike static scenes where single-frame image inpainting may suffice (\eg, SPIN~\cite{mirzaei2023spin} for object removal or NeRFiller~\cite{weber2024nerfiller} for small-hole completion), driving scenarios are characterized by long and noisy camera trajectories, which lead to significant multi-view and temporal inconsistencies.

Traditional 2D inpainting methods struggle under such conditions, as frame-by-frame processing tends to accumulate spatial and temporal artifacts across sequences. Our key insight is that video inpainting, with its ability to model temporal dynamics, inherently enforces consistency across time and improves obstacle removal in cluttered urban environments.

In practice, we first apply a pre-trained semantic segmentation network~\cite{cheng2022masked}to identify dynamic foreground objects (e.g., vehicles, pedestrians) in each frame. These regions are masked and then inpainted using a video inpainting model~\cite{li2025diffueraser}. The resulting clean, temporally coherent images better reflect the true road layout, free from transient occluders.

We integrate these inpainted images into our pipeline in a supervisory role: they serve as training targets for the Gaussian rendering module. Specifically, given an original image $\mathbf{I}$, its inpainted version $\tilde{\mathbf{I}}$, and the rendered image $\hat{\mathbf{I}}$ from our Gaussian representation, we define an inpainting supervision loss. This loss guides the model to reconstruct clean background geometry and appearance, even in regions heavily occluded in the raw input views.

\subsection{Semantic-Aware Color Enhancement}
% \subsection{Multi-Sensor Tight Coupling}
Although the inpainting stage effectively removes transient occlusions, reconstructed road surfaces often exhibit inconsistent color appearance due to varying lighting conditions, shadows, and camera exposure across different views. These variations, if unaddressed, degrade the visual quality and spatial coherence of the final BEV reconstruction. To mitigate this, we introduce a semantic-aware color enhancement module that operates during Gaussian optimization, jointly refining appearance consistency across views while reconstructing the 3D scene geometry.

Our key idea is to leverage semantic correspondence to guide color harmonization. Specifically, we first apply semantic segmentation to identify consistent object categories across multiple views (e.g., road, sidewalk, lane markings). For each semantic region, we establish cross-view correspondences by projecting 3D points into different camera views using known poses. This allows us to collect appearance statistics for the same physical region observed under different conditions.

To make the color enhancement more robust and interpretable, we convert all images from RGB to HSV color space, which decouples luminance (value) from chromaticity (hue and saturation). We then apply per-segment histogram normalization or adaptive tone mapping to align color distributions across views for each semantic class. In practice, we found that enhancing the value (V) and saturation (S) channels leads to more vivid and uniform appearance in the final rendered results, while keeping hue (H) mostly unchanged to preserve semantic identity.

Formally, let $\mathcal{R}_k^c$ denote the $k$-th region of semantic class $c$, and let $\mathbf{I}_v$ be a view where it is visible. For each such region, we compute class-wise HSV statistics $\mu_c^v, \sigma_c^v$, and adjust the value channel of $\mathbf{I}_v$ as:

\begin{equation}
\tilde{V}_v = \frac{\sigma_c^r}{\sigma_c^v} (V_v - \mu_c^v) + \mu_c^r
\end{equation}

where $r$ is a reference view or an averaged target statistic across views. A similar operation is applied to the saturation channel. Finally, enhanced images are used as optional replacements to provide auxiliary supervision in regions affected by illumination artifacts.

While this module is currently integrated as a heuristic adjustment within the Gaussian optimization loop, it opens up potential directions for learnable, end-to-end color harmonization using semantic priors. Future work may incorporate appearance normalization directly into the rendering pipeline or train neural modules to predict lighting-invariant appearance guided by 3D semantics.

\begin{figure*}[!ht] 
    \centering
    \includegraphics[width=1.0\textwidth]{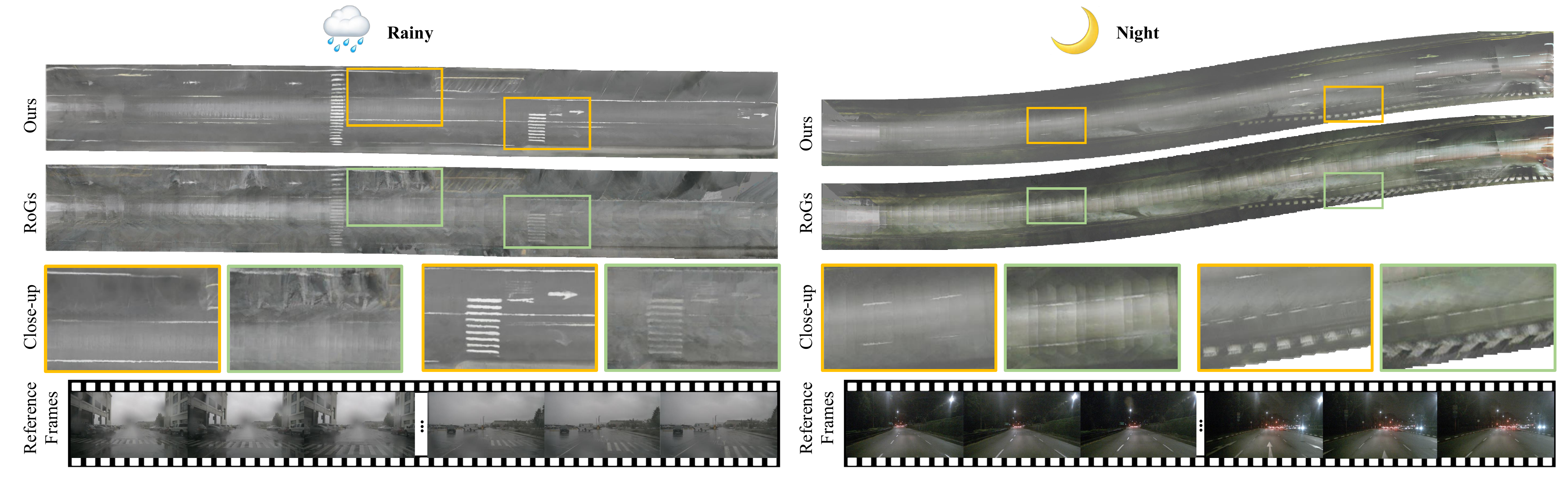}
    % example-grid-100x100pt
    % example-image-a
    \caption{Results under night and rainy conditions. Our method remains robust in challenging environments, effectively removing distracting objects on the road while preserving clear road surface structures and lane markings, demonstrating strong generalization across adverse lighting and weather conditions.} 
    \label{fig:figure6}
\end{figure*}

\subsection{Optimization}
To ensure accurate geometry recovery and consistent surface appearance, we define a composite loss function that supervises both the rendered image quality and the geometric structure of the 2D Gaussian field. The overall loss is defined as:

\begin{equation}
\mathcal{L} = \lambda_{c}\mathcal{L}_{c} + \lambda_{d} \mathcal{L}_{d} + \lambda_{n}\mathcal{L}_{n} + \lambda_{s}\mathcal{L}_{s}
\end{equation}

where $\lambda_{c}, \lambda_{d}, \lambda_{n}, \lambda_{s}$ are the weights for each term. We now describe the individual components:

\noindent\textbf{Photometric Consistency Loss.}
We supervise the rendered image using an L1 photometric loss between the rendered color and the target inpainted image:

\begin{equation}
\mathcal{L}_{c} = \sum_{\bx} \left\| \bc(\bx) - \hat{\bc}(\bx) \right\|_1
\end{equation}

where $\bc(\bx)$ is the predicted color from front-to-back Gaussian blending and $\hat{\bc}(\bx)$ is the corresponding ground truth image (after inpainting and enhancement).

\vspace{0.5em}
\noindent\textbf{Depth Smoothness Loss.}
To encourage smooth surfaces and suppress depth discontinuities between nearby Gaussians, we apply a pairwise depth regularization weighted by their blending strength:

\begin{equation}
\mathcal{L}_{d} = \sum_{i,j}\omega_i\omega_j\,|z_i - z_j|
\end{equation}

where $z_i$ is the depth of the $i$-th Gaussian surfel and $\omega_i$ is its opacity (importance weight) on the current ray.

\noindent\textbf{Normal Consistency Loss.}
To promote locally consistent surface orientation, we penalize deviation of each surfel’s normal $\bn_i$ from the dominant plane normal $\bN$:

\begin{equation}
\mathcal{L}_{n} = \sum_{i} \omega_i \left(1 - \bn_i^\mathrm{T} \bN \right)
\end{equation}

This encourages adjacent surfels to lie on locally planar surfaces, especially in road regions.

\noindent\textbf{Semantic Alignment Loss.}
We further introduce a semantic consistency loss $\mathcal{L}_{s}$ that ensures the Gaussian representation retains semantic boundaries and alignment with the input segmentation. This is implemented by projecting each Gaussian into the segmentation map and minimizing cross-entropy or feature-space distance to its assigned class label.

\section{Experiments}
\label{sec:exp}

\begin{table}[htbp]
\centering
\begin{tabular}{c|cc|cc}
\toprule
 & \multicolumn{2}{c|}{PSNR $\uparrow$} & \multicolumn{2}{c}{RMSE $\downarrow$}  \\
 Method  & Ours & RoGs & Ours & RoGs  \\ \midrule
scene-0008      & 24.12 & 22.47 & 0.039 & 0.057         \\
scene-0026      & 18.98 & 18.19 & 0.126 & 0.152   \\
scene-0098      & 24.47 & 24.40 & 0.036 & 0.036   \\
scene-0121      & 28.47 & 27.15 & 0.014 & 0.019   \\
scene-0125      & 24.46 & 23.85 & 0.035 & 0.041   \\ 
\midrule
Mean            & \textbf{24.10} & 23.21 & \textbf{0.050} & 0.061   \\
\bottomrule
\end{tabular}
\caption{Quantitative results on the nuScenes dataset. We compare PSNR and elevation RMSE across different methods. Our approach achieves significantly higher reconstruction quality and lower geometric error, demonstrating clear improvements in both appearance fidelity and structural accuracy.}
\label{tab:eval_metrics}
\end{table}

\subsection{Experimental Setup}
\paragraph{Datasets.} To validate our method for road surface reconstruction and transient object removal, we conduct experiments on the nuScenes dataset~\cite{caesar2020nuscenes}, a large-scale autonomous driving benchmark comprising 1,000 diverse urban scenes, each 20 seconds long. The dataset features six surround-view cameras and challenging real-world conditions, including dense road markings, dynamic objects, and calibration noise in extrinsics. These characteristics make nuScenes well-suited for evaluating both the visual fidelity and multi-view consistency of our reconstruction framework.

\paragraph{Metrics.} We evaluate our method using two complementary metrics that assess both visual fidelity and geometric accuracy.
Peak Signal-to-Noise Ratio (PSNR) is used to measure the photometric consistency between the rendered view and the ground-truth inpainted image. Higher PSNR indicates better preservation of fine details and overall appearance quality.
To evaluate the accuracy of reconstructed surface geometry, we compute the Root Mean Square Error (RMSE) between the predicted elevation map and the ground-truth depth projected onto the ground plane. This metric reflects how well the model captures fine-grained road surface geometry, particularly in flat and structured areas such as lane markings and curbs.

\paragraph{Baseline methods.}
We primarily compare our approach against RoGs~\cite{feng2024rogs}, a recent method that models road surfaces using meshgrid-based 3D Gaussian splatting. RoGS is specifically designed for BEV-style road reconstruction and has demonstrated strong performance in structured outdoor environments. For a fair comparison, we evaluate our method and RoGs in a lidar-free setting, using the same input images and calibration data.

\begin{figure}[!ht] 
    \centering
    \includegraphics[width=0.48\textwidth]{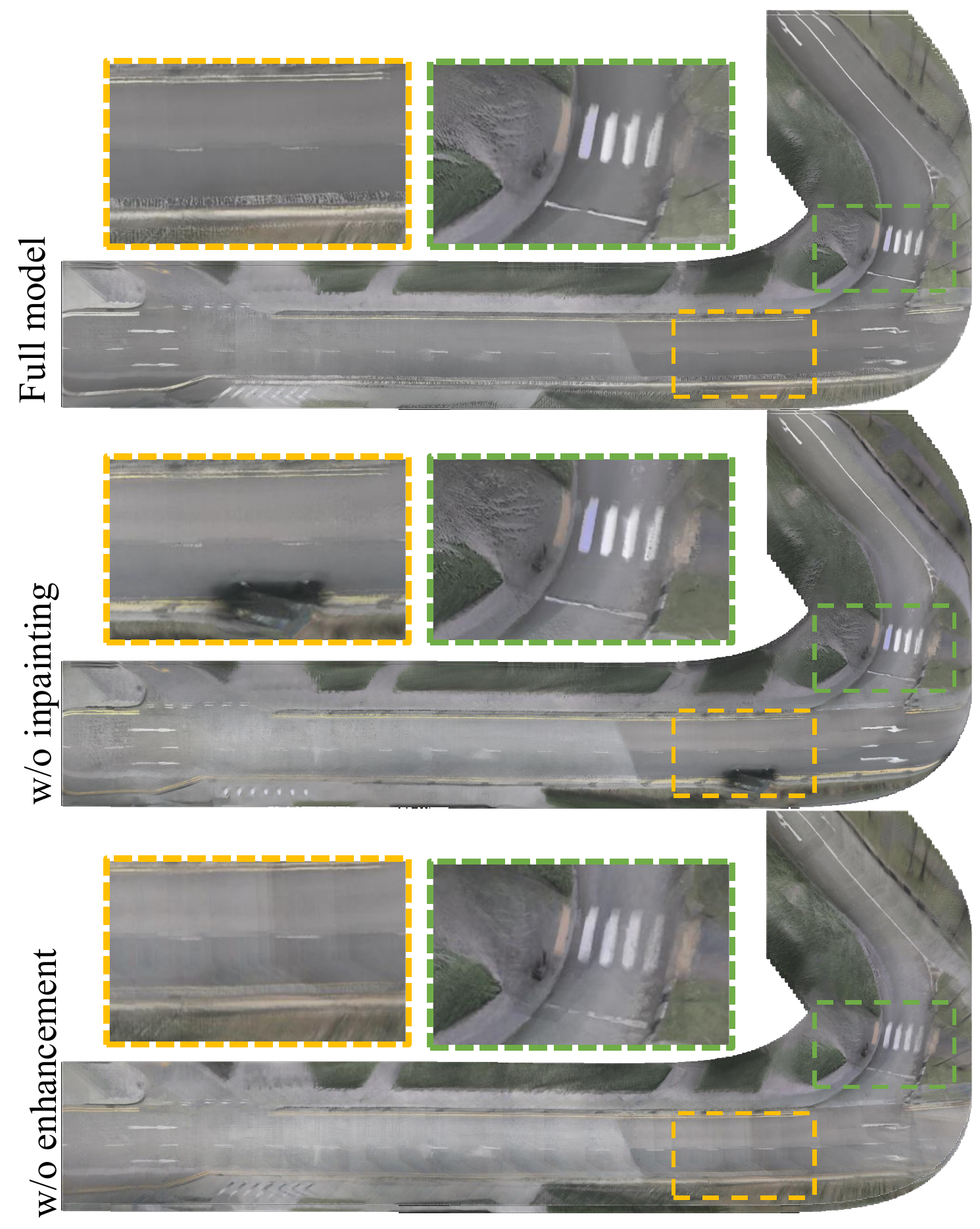}
    % example-grid-100x100pt
    % example-image-a
    \caption{Ablation study on inpainting and color enhancement. Without inpainting, occlusions caused by dynamic and static objects remain in the reconstruction, leading to missing or distorted road surfaces. Without color enhancement, the results exhibit noticeable inconsistencies across views, with uneven illumination and faded lane markings. Our full model achieves the most complete and visually consistent reconstruction.} 
    \label{fig:figure7}
\end{figure}

\subsection{Experimental Results.}
We quantitatively evaluate our method on the nuScenes dataset using two metrics: peak signal-to-noise ratio (PSNR) for appearance fidelity and root mean square error (RMSE) of reconstructed elevation for geometric accuracy. As shown in Table~\ref{tab:eval_metrics}, our method significantly outperforms prior approaches in both metrics. Compared to RoGS and other baselines, we achieve higher PSNR, indicating better preservation of road texture and visual details. At the same time, the lower elevation RMSE demonstrates that our framework more accurately reconstructs the 3D surface geometry, even without relying on lidar supervision. These results validate the effectiveness of our occlusion removal and semantic-guided enhancement modules in improving both visual quality and structural consistency of the reconstruction.

\begin{figure}[!ht] 
    \centering
    \includegraphics[width=0.48\textwidth]{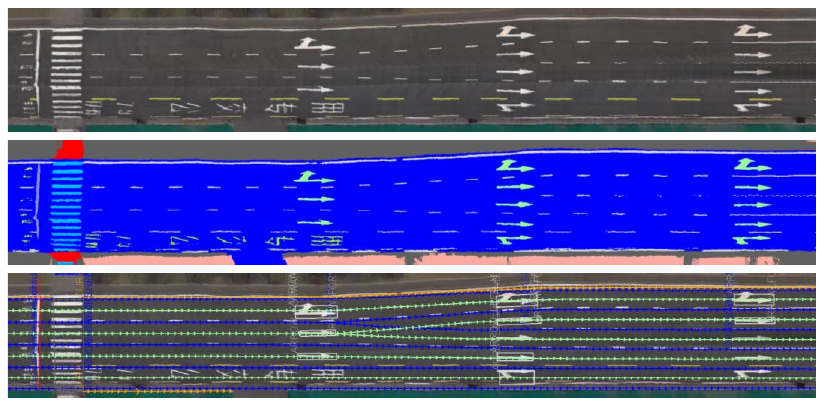}
    % example-grid-100x100pt
    % example-image-a
    \caption{Downstream application enabled by our BEV reconstruction. The high-quality BEV maps produced by our method can directly support downstream tasks. Here, we demonstrate automatic lane marking annotation based on our reconstructed road surfaces, showcasing the utility of our approach for scalable scene understanding and HD map generation.} 
    \label{fig:figure8}
\end{figure}

\paragraph{Downstream Applications.}
We further evaluate the practical utility of our reconstructed BEV surfaces by applying them to automatic lane marking annotation. As shown in Fig.~\ref{fig:figure8}, our method produces clean and well-structured ground-plane representations, enabling robust detection of lane boundaries in complex scenarios. This demonstrates the potential of our method for supporting high-level scene understanding and HD map creation in autonomous driving systems.

\subsection{Ablations and Analysis.}
% We further conduct x ablation experiments on xxx datasets to evaluate xxx. 
To better understand the contribution of each component, we further conduct ablation studies by removing the inpainting and color enhancement modules. Without inpainting, dynamic occlusions such as vehicles and pedestrians remain in the scene, resulting in texture artifacts and increased elevation noise—leading to noticeable drops in both PSNR and elevation accuracy. Without semantic-aware color enhancement, inconsistent lighting across views causes visual artifacts and blurred lane details, which also degrades PSNR. These results demonstrate that both components are critical for achieving high-quality, robust road reconstruction in real-world urban environments.
% \subsection{Limitations}

\section{Conclusion and Discussion}
\label{sec:conclusion}

We proposed a novel framework for large-scale road surface reconstruction that combines semantic-guided video inpainting and color enhancement with 2D Gaussian surfel rendering. Our method effectively removes dynamic and static occlusions, corrects appearance inconsistencies, and produces clean, high-fidelity reconstructions suitable for BEV-based applications. Experiments on nuScenes show significant improvements in both geometry and texture quality. Furthermore, we demonstrate the utility of our outputs for downstream tasks such as automatic lane annotation. Our current method relies on reasonably accurate camera poses for multi-view consistency. In future work, we aim to relax this requirement by exploring joint optimization of pose and reconstruction, or by leveraging robust self-supervised pose refinement strategies.

\bibliography{aaai2026}


\end{document}